\documentclass[letterpaper]{IEEEtran}

\usepackage{cite}
\usepackage{amsmath,amssymb,amsfonts}
\usepackage{gensymb}
\usepackage{textcomp}
\usepackage{xcolor}

\usepackage{algorithm,algpseudocode}

\usepackage{flushend} % split the end evenly
\usepackage{listings}
\DeclareMathOperator*{\argmax}{argmax} % thin space, limits underneath in displays
\usepackage{mathptmx}
\usepackage{graphicx}
\usepackage{textcomp}
\def\BibTeX{{\rm B\kern-.05em{\sc i\kern-.025em b}\kern-.08em
    T\kern-.1667em\lower.7ex\hbox{E}\kern-.125emX}}
\usepackage{makecell}

\usepackage[english]{babel}
\usepackage{amsthm}
\theoremstyle{definition}

\newtheorem{lemma}{Lemma}

\usepackage{caption}
\usepackage{subcaption}
\usepackage{mwe}

\begin{document} 
% Suboptimal Multi-Sensors Deployment for Barrier Coverage in Detecting Poisson-Distributed Target Trajectories

\title{
Improved Approximation of Sensor Network Performance for Seabed Acoustic Sensors
% Enhanced Approximation of Perfect Detection Probability for Poisson Distributed Targets 

% Sensor Location Selection when Detecting Poisson-Distributed Target Trajectories with a Barrier Coverage System
\thanks{*Corresponding author is Mingyu Kim.\\This work was supported in part by the Office of Naval Research under Grant N00014-20-1-2845.}
}

\author{Mingyu Kim$^{1*}$, Daniel J. Stilwell$^1$, Harun Yetkin$^1$, and Jorge Jimenez$^2$ \\ $^1$Bradley Department of Electrical and Computer Engineering,
Virginia Polytechnic Institute and State University, Blacksburg, VA, USA\\
% $^2$ Mechatronics Engineering,  Bart{\i}n University, Turkey \\
$^2$ Johns Hopkins University Applied Physics Laboratory, Laurel, MD, USA\\
$^*$mkim486@vt.edu}

% make the title area
\maketitle

\begin{abstract}

Sensor locations to detect Poisson-distributed targets, such as seabed sensors that detect shipping traffic, can be selected to maximize the so-called void probability, which is the probability of detecting all targets. Because evaluation of void probability is computationally expensive, we propose a new approximation of void probability that can greatly reduce the computational cost of selecting locations for a network of sensors. We build upon prior work that approximates void probability using Jensen’s inequality. Our new approach better accommodates uncertainty in the (Poisson) target model and yields a sharper error bound. The proposed method is evaluated using historical ship traffic data from the Hampton Roads Channel, Virginia, demonstrating a reduction in the approximation error compared to the previous approach. The results validate the effectiveness of the improved approximation for maritime surveillance applications.

\end{abstract}

% \begin{IEEEkeywords}
% 2-D barrier coverage system, sensor placement, log-Gaussian Cox line process, and Poisson-distributed target trajectories
% \end{IEEEkeywords}

\IEEEpeerreviewmaketitle

\section{Introduction}

In this paper, we propose an improved framework for approximating the void probability in underwater sensor networks, specifically for detecting Poisson-distributed targets such as ship traffic or marine life. Void probability, defined as the probability that no targets remain undetected (i.e., probability of perfect detection), is crucial for optimizing sensor placement in scenarios where missing targets is unacceptable. Building on our prior work \cite{kim2023toward}, which proposed a computationally efficient lower-bound approximation of the void probability using Jensen’s inequality, this study incorporates variance into the approximation to enhance accuracy and robustness, especially in high-variance scenarios.

Target arrivals are modeled using a log-Gaussian Cox process (LGCP) \cite{diggle2006spatio,moller1998log}, a Poisson point process where the logarithm of the intensity function follows a Gaussian process. This stochastic model effectively captures spatial variability in target distributions and is widely utilized in fields such as marine mammal surveys \cite{jullum2020estimating}, disease mapping \cite{shirota2017space}, maritime surveillance \cite{kim2023role,kim2024SysCon,kim2023toward} and crime modeling \cite{diggle2005point}. The LGCP framework allows us to represent the target arrival intensity function $\lambda$ as both a mean and a covariance structure, enabling the accurate modeling of spatially distributed events. To estimate the intensity function, we use integrated nested Laplace approximation (INLA) \cite{rue2009approximate, lindgren2015bayesian,bachl2019inlabru}, a computationally efficient Bayesian inference method for latent Gaussian models. This approach facilitates robust estimation of both the mean and variance of the target intensity function, which are crucial for deriving the proposed approximation.

The primary objective of this study is to improve the approximation of the void probability, defined as the probability that no targets remain undetected. 
In our prior work \cite{kim2023toward}, the lower bound of Jensen’s inequality is used to approximate the void probability using a random variable representing the number of undetected target arrivals. While this lower bound is computationally efficient and aids in sensor placement optimization, it overlooks the uncertainty of the intensity function, potentially resulting in inaccurate sensor placements, particularly in scenarios with high variability in target arrival intensities.

To address this limitation, we introduce an approximation based on a second-order Taylor series expansion of the exponential function. The resulting approximation incorporates the variance of the number of undetected targets. This formulation accounts for both the mean and variance of the intensity function, providing a more accurate representation of the void probability. Furthermore, we analyze the relationship between the true void probability and our proposed approximation.
This proposed approximation including the variance potentially improves upon the conventional lower bound. In this paper, we rigorously establish and analyze the gap between the true void probability and the proposed approximation, offering a detailed characterization of this difference.

The main contribution of this paper is the rigorous derivation of a closed-form approximation solution and empirical evaluation of a new approximation to the void probability that accounts for the variance of the underlying intensity function. This approach extends the lower bound derived from Jensen’s inequality in our previous work \cite{kim2023toward}, leading to a more accurate approximation. We prove that the new method yields a maximum error bound that is less than that of the previous approach. We note that our new approximation does not increase computational complexity relative to the earlier approach.

% The main contribution of this paper is a rigorous derivation and empirical evaluation of a new approximation to void probability that incorporates the variance of the corresponding intensity function, extending the lower bound derived from Jensen’s inequality in our prior work \cite{kim2023toward}, for a more accurate approximation. We prove that our new approximation provides less or equal maximum error bound that that of the previous method.c We note that our new approximation does not increase computational complexity relative to the earlier approach.

To evaluate the proposed approach, we use historical ship arrival data from the Hampton Roads Channel, Virginia, USA. This dataset, provided by the Bureau of Ocean Energy Management (BOEM) and National Oceanic and Atmospheric Administration (NOAA) \cite{marinecadastre.gov}, is shown in Fig. \ref{fig:1}. Using this ship traffic data, we estimate the intensity function of ship arrivals using INLA\cite{rue2009approximate, lindgren2015bayesian,bachl2019inlabru}, which computes the mean and variance of the ship arrival intensity function. 

% The paper is organized as follows. In Section~\ref{sec:ProblemFormulation}, we introduce our proposed approximation of the void probability and establish an inequality demonstrating its improved maximum error bound of the approximation to the void probability. Section~\ref{sec:GapAnalysis} analyzes the difference between the original objective function (void probability) and the proposed approximation. In Section~\ref{sec:NumericalResults}, we present numerical experiments to evaluate the efficiency of our method. The appendix provides proof of the proposed inequality, comparing our method to the lower bound derived from Jensen's inequality. 

The paper is organized as follows. In Section~\ref{sec:ProblemFormulation}, we introduce our proposed approximation of the void probability and show 
 the derivation of the closed-form approximation solution. Section~\ref{sec:GapAnalysis} analyzes the difference between the original objective function (i.e., void probability) and the proposed approximation, and establishes an inequality demonstrating its improved maximum error bound of the approximation to the void probability. In Section~\ref{sec:NumericalResults}, we present numerical experiments to evaluate the efficiency of our method. The appendix provides proof of the proposed inequality, comparing our method to the lower bound derived from Jensen's inequality. 

% Numerical simulations demonstrate that the variance-inclusive framework improves the approximation accuracy. By reducing the Jensen gap, the proposed method ensures a closer alignment between the true void probability and its approximation. The greedy algorithm, employed for sensor placement, leverages the submodular and monotonic properties of the objective function to identify near-optimal sensor locations efficiently. This algorithm offers computational advantages and scalability, making it suitable for real-time applications in dynamic underwater environments.

% This study strengthens the theoretical foundation of underwater sensor networks by improving the approximation of void probability for Poisson-distributed targets. By incorporating variance into the detection model, the proposed framework achieves a more accurate approximation without increasing computational time. 

\section{Problem Formulation: \\ Void probability approximation incorporating both mean and variance}\label{sec:ProblemFormulation}

% Our original objective function from \cite{kim2023optimal} is the void probability by a sensor network $\mathbf{a}$ within a bounded domain $\Psi$
% \begin{align}
% \mathbb{E}_{\lambda}\left[e^{-X(\mathbf{a},\Psi)} \right]
% \label{eq: original objective fn}
% \end{align} 
% where $X(\mathbf{a},\Psi)$ represents the number of undetected targets by a sensor network $\mathbf{a}=\{a_1,a_2,...,a_n\}$ $\forall a_i \in \Psi$
% \begin{align}    X(\mathbf{a},\Psi)=\int_{\Psi} \lambda(s) \prod_{i=1}^{n} (1-\gamma(s,a_i)) ds \label{eq: num undetected targets}
% \end{align}
%   where $\lambda(s)$ represents the intensity function of target arrivals at $s\in \Psi$, which is stochastic
%   \begin{align*}
%       \log(\lambda(s))\sim GP(\mu_{\lambda}(s),\Sigma_{\lambda}(s,s'))
%   \end{align*}
%   where $\mu_{\lambda}(s),\Sigma_{\lambda}(s,s'))$ are the mean and covariance of $\lambda(s)$.

Our original objective function, introduced in \cite{kim2023toward}, represents the void probability of a sensor network $\mathbf{a}$ within a bounded domain $\Psi$
\begin{align} \mathbb{E}_{\lambda}\left[e^{-X(\mathbf{a},\Psi)} \right] \label{eq: original objective fn} \end{align}
\noindent where $X(\mathbf{a},\Psi)$ denotes the number of undetected targets given a sensor network $\mathbf{a} = \{a_1,a_2,\dots,a_n\}$ with sensor locations $a_i \in \Psi$
\begin{align} 
    X(\mathbf{a},\Psi) = \frac{T}{T_c}\int_{\Psi} \lambda(s) \prod_{i=1}^{n} (1-\gamma(s,a_i)) ds \label{eq: num undetected targets} 
\end{align}

\noindent Here, \( T \) and \( T_c \) denote the time periods of interest and the collected historical data, respectively, while \( \lambda(s) \) represents the intensity function of target arrivals at location \( s \in \Psi \). The intensity function \( \lambda(s) \) is modeled as a stochastic process, specifically a log-Gaussian Cox process (LGCP)

\begin{align*} \log(\lambda(s)) \sim GP(\mu_{\lambda}(s),\Sigma_{\lambda}(s,s')), \end{align*}

\noindent where $\mu_{\lambda}(s)$ and $\Sigma_{\lambda}(s,s')$ denote the mean and covariance of $\lambda(s)$, respectively. For clarity and conciseness, we denote $X(\mathbf{a},\Psi)$ as $X$ for the remainder of the paper.

  Probability of detection of target at $s$ by a sensor $i$ at $a_i$ is denoted as $\gamma(s,a_i)$. Correspondingly, the probability of failing detection by a set of sensors $\mathbf{a}$ is 
 \begin{align*}
     \prod_{i=1}^{n} (1-\gamma(s,a_i))
 \end{align*}
 where $n$ is the number of sensors and $a_i \in \Psi$ represents the location of the sensor $i$.
\subsection{Proposed void probability approximation}

 We account for uncertainty in the intensity function $\lambda$ in (2) using the variance of $\lambda$.  Approximating $\mathbb{E}_{\lambda}[e^{-X}]$ using a second-order Taylor series around $\mathbb{E}_{\lambda}[X]$ yields a closed-form approximation in terms of $\mathbb{E}_{\lambda}[X]$ and $\sigma_{X}^2$
\begin{align}
    \mathbb{E}_{\lambda}[e^{-X}]&\approx  e^{-\mathbb{E}_{\lambda}[X]} \left(1 + \frac{1}{2}\sigma_{X}^2 \right)\label{eq: approx solution}
\end{align} 
where  $\sigma_{X}^2=\mathbb{E}_{\lambda}[X^2]-\mathbb{E}_{\lambda}[X]^2$ represents the variance of $X$. To compute the variance $\sigma_{X}^2$, we separately evaluate these two components $\mathbb{E}_{\lambda}[X^2]$ and $\mathbb{E}_{\lambda}[X]^2$. For the first component, $\mathbb{E}_{\lambda}[X]^2$, we can take the expected value of $\lambda$ in \eqref{eq: num undetected targets} and then square the entire equation
\begin{align*}
    \mathbb{E}_{\lambda}[X]^2=\left(\frac{T}{T_c}\right)^2\int_{\Psi} \mathbb{E}_{\lambda}[\lambda(s)]^2 \prod_{i=1}^{n} (1-\gamma(s,a_i))^2 ds
\end{align*}

\noindent Next, to compute the second component, $\mathbb{E}_{\lambda}[X^2]$, we evaluate $X^2$ and then take the expected value with respect to $\lambda$
\begin{align*} 
\mathbb{E}_{\lambda}[X^2] = \left(\frac{T}{T_c}\right)^2\int_{\Psi} \big(\sigma_{\lambda}(s)^2 + \mathbb{E}_{\lambda}[\lambda(s)]^2\big) \prod_{i=1}^{n} (1-\gamma(s,a_i))^2 ds\end{align*}
This expression is simplified using the definition of variance, where $\sigma_{\lambda}(s)^2$ represents the variance of $\lambda(s)$. Therefore, the variance of the number of undetected targets $\sigma_X^2$ is
\begin{align*}
    \sigma_{X}^2 
    = \left(\frac{T}{T_c}\right)^2\int_{\Psi} \sigma_{\lambda}(s)^2  \prod_{i=1}^{n} (1-\gamma(s,a_i))^2ds
\end{align*}

\noindent We note that $\sigma_{X}^2$ is always positive so that the proposed approximation is always greater than our previous approximation $ e^{-\mathbb{E}_{\lambda}[X]}$, which is the lower bound of Jensen's inequality of $\eqref{eq: original objective fn}$.

\subsection{Searching for Optimal Sensor Locations}

When choosing sensor locations, we seek to maximize void probability, which is equivalent to minimizing the number of targets that are not detected. Because void probability is computationally expensive to compute, we instead choose sensor locations that maximize our proposed approximation.  That is, 

\begin{equation}
    \mathbf{a}^\star = \argmax_{\mathbf{a} \subset \mathbf{A}} e^{-\mathbb{E}_{\lambda}[X]} \left(1 + \frac{1}{2}\sigma_{X}^2 \right)
    \label{eq:optimalSensorLocations}
\end{equation}

% \noindent To ensure computational efficiency in real-time applications, we employ a greedy selection strategy that iteratively searches for a sensor location that maximizes the objective function in \eqref{eq:optimalSensorLocations}. This approach provides a practical balance between solution quality and computational feasibility, making it particularly suitable for dynamic sensor deployment scenarios. Moreover, as proved in \cite{kim2023toward}, the greedy selection to search for the sensor locations  applied to the lower bound derived from Jensen's inequality guarantees at least \(63.2\%\) optimality with respect to the lower bound. Building on this result, our proposed approximation with the greedy selection to search is expected to provide an even tighter approximation to the true optimal solution.

\noindent To ensure computational efficiency in real-time applications, we employ a greedy selection strategy that iteratively identifies sensor locations maximizing the objective function in \eqref{eq:optimalSensorLocations}. This approach provides a practical balance between solution quality and computational feasibility, making it particularly suitable for dynamic sensor deployment scenarios. Moreover, as shown in \cite{kim2023toward}, applying the greedy selection to the lower bound derived from Jensen's inequality guarantees at least \(63.2\%\) optimality with respect to the lower bound.

\section{Gap analysis: true void probability and our proposed void probability approximation} \label{sec:GapAnalysis}

% \section{Numerical Analysis of the Jensen Gap}

In this section, we numerically analyze the gap between the original objective function and the proposed approximated objective function. In \cite{kim2023toward}, we use Jensen's inequality to derive a lower bound $e^{-\mathbb{E}_{\lambda}[X]}$ of the objective function $\mathbb{E}_{\lambda}[e^{-X}]$.  The lower bound leads to a computationally easier maximization problem than the original objective function.  The difference between the objective function and its lower bound is known as Jensen's gap, defined in our case as

\begin{align}
    J = \mathbb{E}_{\lambda}[e^{-X}] - e^{-\mathbb{E}_{\lambda}[X]} \label{eq: Jensen gap}
\end{align}

\noindent Using Theorem 2 and Equation (12) from \cite{kim2023toward}, the bound of Jensen gap is expressed 
\begin{align*}
    J_{low} \leq J \leq J_{up}
\end{align*}
where
\begin{align}
    J_{up} &= \sup_{X\in[0,\infty)}  
    \frac{\sigma_{X}^2\left(e^{-X}-e^{-\mu_{X}}+Xe^{-\mu_{X}}-\mu_{X}e^{-\mu_{X}}\right)}{(X-\mu_{X})^2} \label{Jensen gap up} \\
     J_{low} &= \inf_{X\in[0,\infty)}  
    \frac{\sigma_{X}^2\left(e^{-X}-e^{-\mu_{X}}+Xe^{-\mu_{X}}-\mu_{X}e^{-\mu_{X}}\right)}{(X-\mu_{X})^2} \label{Jensen gap low}
\end{align}
 We define $\mu_{X} = \mathbb{E}_{\lambda}[X]$ for simplicity and conciseness. In \cite{kim2023toward}, Appendix B shows that 
 $\sigma_{X}^2\left(e^{-X}-e^{-\mu_{X}}+Xe^{-\mu_{X}}-\mu_{X}e^{-\mu_{X}}\right)/(X-\mu_{X})^2 $ is monotonic-decreasing with respect to $X$. Therefore, the bound of $J$ is maximized and minimized when $X = 0$ and $X \rightarrow \infty$, respectively. That is
\begin{align}
    0 \leq J \leq \frac{\sigma_{X}^2\left(1-e^{-\mu_{X}}-\mu_{X}e^{-\mu_{X}}\right)}{\mu_{X}^2} \label{eq: Jensen gap bound}
\end{align}

\subsection{Bound of Gap between the Void Probability and the Proposed Void Probability Approximation}

%%%%%%%%%%%%%%%%%%%%%%%%%%%%%%%%%%%%%%%%%%%%%%%%%%%%
\begin{figure}[t!] % figure 1
\centering
\includegraphics[scale=0.65]{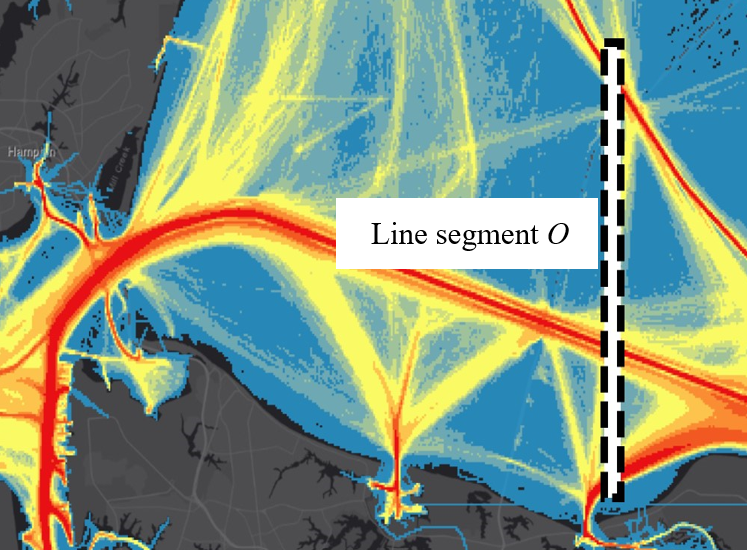}
\caption{Heatmap of ship traffic data with a line segment $O$ near Hampton Roads Channel, Virginia, USA \cite{marinecadastre.gov}}
\label{fig:1}
\end{figure}
%%%%%%%%%%%%%%%%%%%%%%%%%%%%%%%%%%%%%%%%%%%%%%%%%%%%
Compared to the lower bound of Jensen's inequality, given by $e^{-\mu_{X}}$, our proposed approximation in \eqref{eq: approx solution} introduces an additional term, $-\frac{1}{2} e^{-\mu_{X}} \sigma_{X}^2$, accounting for variance so that the new gap $\Tilde{J}$ becomes

\begin{align} \Tilde{J} &= \mathbb{E}_{\lambda}\left[e^{-X}\right] - e^{-\mu_{X}} \left(1 + \frac{1}{2} \sigma_{X}^2 \right) \label{eq: new gap}\\ &= J - \frac{1}{2} e^{-\mu_{X}} \sigma_{X}^2 \nonumber
\end{align}
Using this additional term and \eqref{eq: Jensen gap bound}, we get a corresponding bound of the gap for the error of our proposed approximation
\begin{align}
    \Tilde{J}_{low} \leq \Tilde{J} \leq \Tilde{J}_{up}\label{eq: new gap bound}
\end{align}
where
\begin{align}
    \Tilde{J}_{low}&=  -\frac{1}{2} e^{-\mu_{X}} \sigma_{X}^2 \label{eq: new bound low} \\ 
    \Tilde{J}_{up} &=\frac{\sigma_{X}^2\left(1-e^{-\mu_{X}}-\mu_{X}e^{-\mu_{X}}-\frac{1}{2}\mu_X^2 e^{-\mu_{X}}\right)}{\mu_{X}^2}\label{eq: new bound up}
\end{align}
 We note that the bound of the new gap in \eqref{eq: new gap bound} is determined by the mean and variance of the random variable $X$, which represents the number of undetected target arrivals by a sensor network found by solving the optimization problem in \eqref{eq:optimalSensorLocations}.
%%%%%%%%%%%%%%%%%%%%%%%%%%%%%%%%%%%%%%%%%%%%%%%%%%%%
\begin{figure}[t!] % figure 2
\centering
\includegraphics[scale=0.28]{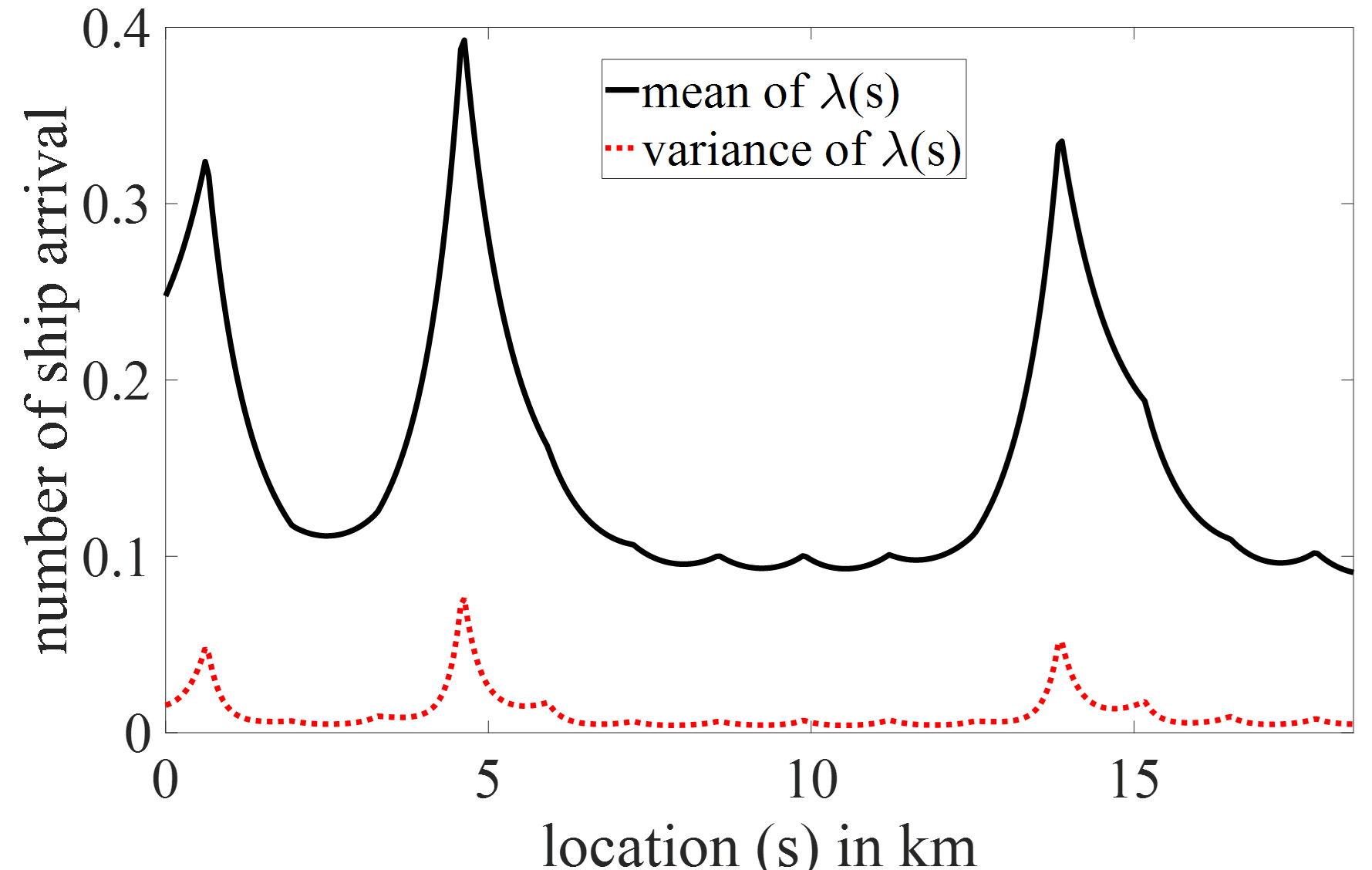}
\caption{Estimated mean and variance of the ship arrival intensity functions along the line segment $O$ in Fig. \ref{fig:1}}
\label{fig:2}
\end{figure}
%%%%%%%%%%%%%%%%%%%%%%%%%%%%%%%%%%%%%%%%%%%%%%%%%%%%

% \begin{lemma}
% Let $\Omega$ be a positive random variable with the corresponding expectation $\mathbb{E}_{\lambda}[\Omega]$ and variance $\sigma_{\Omega}^2$. Then, the following inequality holds
% \begin{align}
%     \max(|J(\Omega)|) \geq \max(|\Tilde{J(\Omega)}|) 
% \label{eq: proposed inequality}
% \end{align}
% where $\Tilde{J}(\Omega)$ and $J(\Omega)$ represent the approximation errors to the void probability by $e^{-\mathbb{E}_{\lambda}[\Omega]} \left(1 + \frac{1}{2}\sigma_{\Omega}^2 \right)$ and  $e^{-\mathbb{E}_{\lambda}[\Omega]} $, respectively
% \begin{align*}
%     \Tilde{J}(\Omega)&=    \mathbb{E}_{\lambda}\left[e^{-\Omega}\right] -  e^{-\mathbb{E}_{\lambda}[X]} \left(1 + \frac{1}{2}\sigma_{\Omega}^2 \right) \\
%     J(\Omega) &=    \mathbb{E}_{\lambda}\left[e^{-\Omega}\right] - e^{-\mathbb{E}_{\lambda}[\Omega]} 
% \end{align*} 
% \end{lemma} 

\begin{lemma}
Let $\Omega$ be a positive random variable with expectation $\mathbb{E}_{\lambda}[\Omega]>0$ and variance $\sigma_{\Omega}^2>0$. Then, the following inequality holds
\begin{align}
    \max\left(|J(\Omega)|\right) > \max\left(|\Tilde{J}(\Omega)|\right),
\label{eq: proposed inequality}
\end{align}
where $J(\Omega)$ and $\Tilde{J}(\Omega)$ denote the approximation errors in estimating the void probability $\mathbb{E}_{\lambda}[e^{-\Omega}]$ using $e^{-\mathbb{E}_{\lambda}[\Omega]}$ and $e^{-\mathbb{E}_{\lambda}[\Omega]} \left(1 + \frac{1}{2}\sigma_{\Omega}^2 \right)$, respectively
\begin{align*}
    J(\Omega) &= \mathbb{E}_{\lambda}\left[e^{-\Omega}\right] - e^{-\mathbb{E}_{\lambda}[\Omega]}, \\
    \Tilde{J}(\Omega) &= \mathbb{E}_{\lambda}\left[e^{-\Omega}\right] - e^{-\mathbb{E}_{\lambda}[\Omega]} \left(1 + \frac{1}{2}\sigma_{\Omega}^2 \right).
\end{align*}
\end{lemma}

In Lemma 1, we show that the proposed approximation method $e^{-\mathbb{E}_{\lambda}[X]} \left(1 + \frac{1}{2}\sigma_{X}^2 \right)$ in \eqref{eq: approx solution} provides a tighter maximum approximation error bound compared to $e^{-\mathbb{E}_{\lambda}[X]}$. The proof of Lemma 1 is provided in Appendix A.  Building on this result, our proposed approximation, based on greedy selection, is expected to achieve a tighter approximation to the true void probability.

\section{Numerical Result}\label{sec:NumericalResults}

In this section, we assess the effectiveness of our proposed approximation of the void probability  through a numerical experiment using ship traffic data from \cite{marinecadastre.gov}, as shown in Fig.\ref{fig:1}. Using the ship traffic data, we estimate its intensity function and focus on searching for the optimal seabed sensor locations for detecting Poisson-distributed ship arrivals. Specifically, we analyze ship traffic data from March 2022 along an 18.5 km long line segment $O$ (see Fig. \ref{fig:1}) of the Hampton Roads Channel, which contains three major high-traffic zones. The heatmap in Fig.\ref{fig:1} visually represents traffic intensity, where red indicates high traffic, yellow represents low traffic, and blue denotes no traffic. 

While this experiment is conducted in a one-dimensional setting, our framework can be extended to higher-dimensional cases, where the number of undetected ships is modeled as a log-Gaussian Cox process. By incorporating uncertainty into the void probability approximation, our method more accurately captures the estimation of the uncertain ship arrival intensity function, resulting in a tighter maximum error bound in the representation of detection performance.
\subsection{Estimating Mean and Variance of Ship Arrival Intensity function}

To estimate the mean and variance of the intensity function, we utilize the \texttt{inlabru} package in R~\cite{bachl2019inlabru}, which is built on top of the R-INLA framework~\cite{lindgren2015bayesian}. The model is based on a Gaussian process with zero mean and a Matérn covariance function, given by
\begin{equation*}
    k(s,s^\prime)= \sigma_u^2 \ \frac{2^{1-\zeta}}{\Gamma(\zeta)} \ (\kappa ||s - s^\prime || )^{\zeta} \ K_{\zeta} \ (\kappa ||s - s^\prime ||)
\end{equation*}

\noindent where $s$ and $s^\prime$ represent spatial locations within the domain, $\sigma_u$ is the variance, and $\zeta > 0$ is the smoothness parameter. The term $\kappa = \sqrt{8\zeta}/\beta > 0$ denotes the scale parameter, while $|| \cdot ||$ represents the Euclidean distance. Additionally, $K_\zeta$ is the modified Bessel function of the second kind, and $\beta$ serves as the spatial range parameter (see~\cite{rpackagedocumentation_2019} for additional details).

%%%%%%%%%%%%%%%%%%%%%%%%%%%%%%%%%%%%%%%%%%%%%%%%%%%%
\begin{figure}[t!] % fig3
\centering
\includegraphics[scale=0.28]{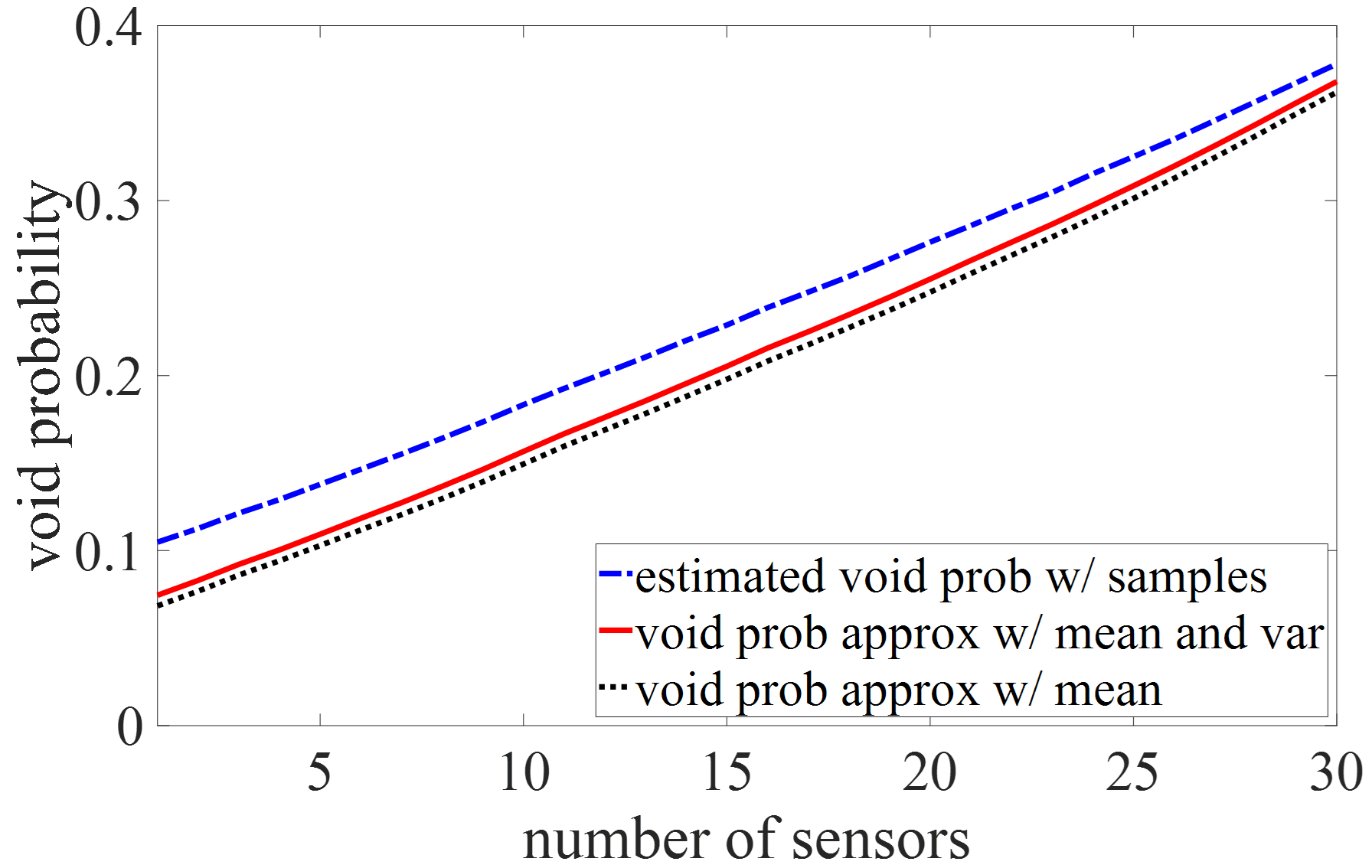}
\caption{Estimated void probability (blue dashed curve) and void probability approximations (red and black dotted curves) with greedily selected sensor locations}
\label{fig:3}
\end{figure}
%%%%%%%%%%%%%%%%%%%%%%%%%%%%%%%%%%%%%%%%%%%%%%%%%%%%

\noindent For numerical simulations, we set the parameters as follows $\zeta = 1.5$, while $\beta$ and $\sigma_u$ are selected based on the probability thresholds $P(\beta < \beta_0 = 150) = 0.75$ and $P(\sigma_u > \sigma_{u0} = 0.1) = 0.75$. Using these parameter values, along with the defined covariance function and historical ship traffic data, we apply INLA to estimate both the mean and variance of the ship arrival intensity functions. Fig.\ref{fig:2} shows the estimated mean and variance of the ship arrival intensity function along the line segment $O$ from Fig.\ref{fig:1}, which are computed when we use INLA.

\subsection{1-D Sensor Model}

To represent the probability of sensor $i$ detecting a ship arrival, we adopt the following sensor detection model

\begin{equation}
    \gamma(s, a_i) = \rho e^{-(a_i - s)^2 / \sigma_l}
    \label{eq:probabilityOfDetection}
\end{equation}

\noindent where $0 \leq \rho \leq 1$ denotes the maximum detection probability, and $\sigma_l$ represents the length scale parameter, controlling the spatial sensitivity of the sensor. For numerical simulations, we set $\rho = 0.95$ and $\sigma_l = 0.05$, ensuring that the detection probability significantly decays as the distance from the sensor increases. This configuration allows the sensor to detect targets within an approximate range of 0.5 km.

\subsection{Sensor Placement Algorithm for Maximizing the Improved Void Probability Approximation}

% Our objective is to determine a set of sensor locations that maximize the proposed void probability approximation in \eqref{eq: approx solution}. For the possible sensor locations are discretized between 0 km to 18.5 km with a 50 m interval. Given the computational complexity of a brute force search, we employ a greedy selection method to obtain a computationally efficient solution. Specifically, we iteratively select sensor locations $\hat{\mathbf{a}} = \{\hat{a}_1, \hat{a}_2, \dots, \hat{a}_n\}$ that maximize \eqref{eq: approx solution} at each sensor. 

Our objective is to determine the optimal set of sensor locations that maximize the proposed void probability approximation in \eqref{eq: approx solution}. The potential sensor locations are discretized along a 0 km to 18.5 km range with a 50 m interval. Due to the computational burden of a brute force search to find the optimal sensor locations, we adopt a greedy selection approach to achieve a computationally efficient solution. In this method, sensor locations $\hat{\mathbf{a}} = \{\hat{a}_1, \hat{a}_2, \dots, \hat{a}_n\}$ are iteratively chosen to maximize \eqref{eq: approx solution} at each step, ensuring computational feasibility.

% Using the sensor locations obtained through greedy selection applied to \eqref{eq:optimalSensorLocations}, Fig.~\ref{fig:3} illustrates the estimated void probability (blue dashed curve), the void probability approximation from our proposed method (red curve), and the lower bound given by Jensen's inequality (dotted black curve) as the number of sensors increases from 1 to 30. The estimated void probability is computed using a large number, $M$ (in our example, $M=20,000$), of sampled intensity functions $\Tilde{\lambda}_j$ and the selected sensor locations $\hat{\mathbf{a}}$:
%%%%%%%%%%%%%%%%%%%%%%%%%%%%%%%%%%%%%%%%%%%%%%%%%%%%
\begin{figure}[t!] % fig4
\centering
\includegraphics[scale=0.27]{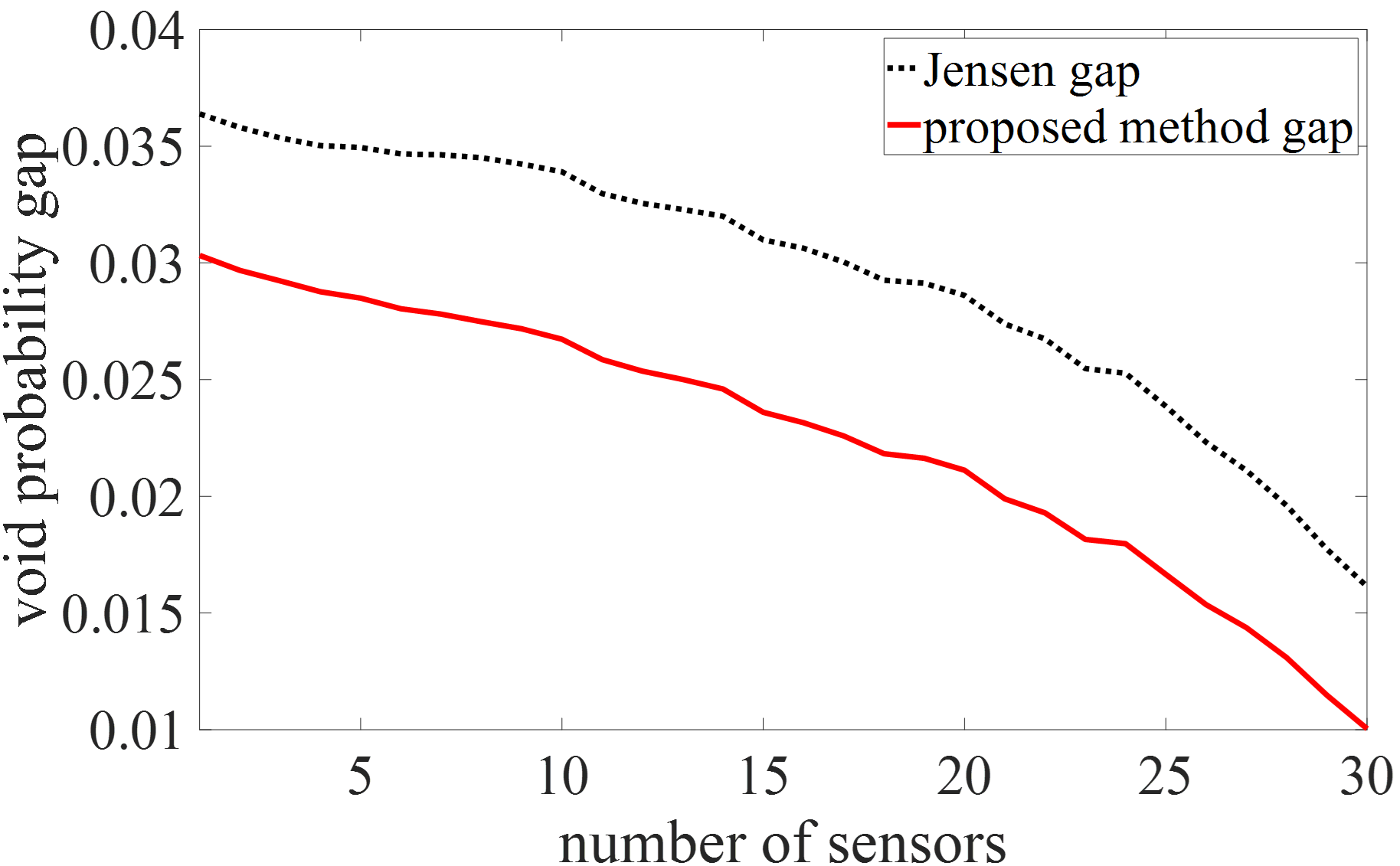}
\caption{Void probability difference from the estimated void probability: 1) the lower bound of Jensen's inequality (black dotted curve) and 2) proposed method  (red curve)}
\label{fig:4}
\end{figure}
%%%%%%%%%%%%%%%%%%%%%%%%%%%%%%%%%%%%%%%%%%%%%%%%%%%%
Fig.~\ref{fig:3} shows the estimated void probability (blue dashed curve), the void probability approximation from our proposed method (red curve), and the lower bound derived from Jensen's inequality (dotted black curve) as the number of sensors increases from 1 to 30. These results are based on sensor locations obtained using the greedy selection method applied to \eqref{eq:optimalSensorLocations}. The estimated void probability is calculated by averaging over a large set of Monte Carlo sampled intensity functions, $\Tilde{\lambda}_j$, with $M = 20,000$ samples in this study, using the selected sensor locations $\hat{\mathbf{a}}$

\begin{align*}
    \frac{1}{M} \sum_{j}^{M} \exp\left(-\left(\frac{T}{T_c}\right)^2
    \int_{\Psi} \Tilde{\lambda}_j(s) \prod_{i=1}^{n} (1-\gamma(s,\hat{a}_i)) ds \right).
\end{align*}
 Furthermore, Fig.~\ref{fig:4} quantifies the deviation of the void probability approximations $e^{-\mathbb{E}_{\lambda}[X]}$ and $e^{-\mathbb{E}_{\lambda}[X]} \left(1 + \frac{1}{2} \sigma_{X}^2 \right)$ from the estimated void probability as sensors are greedily placed from 1 to 30, consistent with the results in Fig.~\ref{fig:3}. The black and red curves represent the gaps $\mathbb{E}_{\lambda}[e^{-X}] - e^{-\mathbb{E}_{\lambda}[X]}$ and $\mathbb{E}_{\lambda}[e^{-X}] - e^{-\mathbb{E}_{\lambda}[X]} \left(1 + \frac{1}{2} \sigma_{X}^2 \right)$, respectively. Compared to the lower bound derived from Jensen's inequality, our proposed approximation reduces the average gap by approximately $24.68\%$, demonstrating its improved accuracy in capturing the true void probability.

\section{Conclusion}

This paper presents an improved approximation method for estimating the void probability in underwater sensor networks used to detect Poisson-distributed targets. By incorporating the variance of the number of undetected targets into the approximation, we extend the lower bound derived from Jensen’s inequality, leading to a more accurate estimation of the void probability. We mathematically establish this improvement and analyze the gap between the true void probability and our proposed approximation. To efficiently determine optimal sensor placements, we employ a greedy selection approach that balances computational efficiency with solution quality. Numerical experiments using historical ship traffic data from the Hampton Roads Channel validate the effectiveness of our method, demonstrating a reduction in the difference from the void probability compared to the previous approximation, a lower bound of Jensen's inequality. The proposed framework enhances the accuracy of estimation of void probability without increasing computational complexity, making it suitable for real-time applications.

\section*{Appendix}

\subsection{Proof of Lemma 1 (positivity of $  \max(|J|) - \max(|\Tilde{J}|)$)}

In this section, we prove the inequality \eqref{eq: proposed inequality} by showing that
\begin{align}
    \max(|J|) - \max(|\Tilde{J}|) > 0 \label{eq: ineq proof}
\end{align}
To evaluate $\max(|J|)$, which represents the maximum approximation deviation in magnitude from $\mathbb{E}_{\lambda}\left[e^{-X}\right]$, we use Jensen's inequality, which implies that $J$ is always positive
\begin{align*}
    J = \mathbb{E}_{\lambda}\left[e^{-X}\right] - e^{-\mathbb{E}_{\lambda}[X]} \geq 0
\end{align*}
Therefore, the upper bound on $|J|$ is given by the known Jensen gap bound
\begin{align}
    0 \leq |J| \leq \frac{\sigma_{X}^2\left(1 - e^{-\mu_{X}} - \mu_{X} e^{-\mu_{X}}\right)}{\mu_{X}^2} \label{eq: Jensen gap bound}
\end{align}
where $\mu_X := \mathbb{E}_{\lambda}[X]$. However, for $\Tilde{J}$, the error term may be positive or negative depending on the values of $\mu_X$ and $\sigma_X^2$ as shown in \eqref{eq: new gap bound}, so we must consider the magnitudes of its upper and lower bounds. In particular, we need to determine whether $\Tilde{J}_{\text{up}}$ or $\Tilde{J}_{\text{low}}$ has the greater absolute value as $\mu_X$ varies from $0$ to $\infty$. Since $\sigma_X^2$ is positive, it does not affect the direction of the inequality.

To prove \eqref{eq: ineq proof}, it suffices to show the following two inequalities hold
\begin{align}
    J_{\text{up}} >\Tilde{J}_{\text{up}}, \label{eq: cond1} \\
    J_{\text{up}} >|\Tilde{J}_{\text{low}}| \label{eq: cond2}
\end{align}

\paragraph{Verification of \eqref{eq: cond1}}  
This inequality holds trivially since
\begin{align*}
    \Tilde{J}_{\text{up}} = J_{\text{up}} - \frac{1}{2}e^{-\mu_X} \sigma_X^2,
\end{align*}
and both $\sigma_X^2$ and $e^{-\mu_X}$ are positive Hence, $\Tilde{J}_{\text{up}} < J_{\text{up}}$.

\paragraph{Verification of \eqref{eq: cond2}}  
We show that $J_{\text{up}} > |\Tilde{J}_{\text{low}}|$ by rewriting the expression using bounds 
\begin{align}
    J_{\text{up}} - |\Tilde{J}_{\text{low}}| 
    &= \frac{\sigma_X^2 \left(1 - e^{-\mu_X} - \mu_X e^{-\mu_X} - \frac{1}{2} \mu_X^2 e^{-\mu_X} \right)}{\mu_X^2} \label{eq: Tilde J_up} \\
    &= \Tilde{J}_{up} \nonumber
\end{align}
and check if $\Tilde{J}_{up} > 0$. Since $\sigma_X^2, \mu_X^2$ are greater than zero, letting 
\[
\xi(\mu_X) = 1 - e^{-\mu_X} - \mu_X e^{-\mu_X} - \frac{1}{2} \mu_X^2 e^{-\mu_X}
\]
we investigate whether $\xi(\mu_X) > 0$ for all $\mu_X \in [0, \infty)$.We observe that $\xi(\mu_X)$ is monotonically increasing for $\mu_X > 0$, since its derivative is
\begin{align*}
    \frac{d\xi(\mu_X)}{d\mu_X} = \frac{1}{2} \mu_X^2 e^{-\mu_X} > 0
\end{align*}
Here, both $\mu_X^2$ and $e^{-\mu_X}$ are positive for $\mu_X > 0$. Therefore, as $\mu_X \to 0^{+}$, the minimum value of $\xi(\mu_X)$ goes to $0^{+}$.  Therefore, \eqref{eq: cond2} holds for all $\mu_X > 0$, and we conclude that the inequality \eqref{eq: ineq proof} always holds \qedsymbol.

\bibliographystyle{IEEEtran}
\bibliography{main.bib}

% Generated by IEEEtran.bst, version: 1.14 (2015/08/26)
\begin{thebibliography}{10}
\providecommand{\url}[1]{#1}
\csname url@samestyle\endcsname
\providecommand{\newblock}{\relax}
\providecommand{\bibinfo}[2]{#2}
\providecommand{\BIBentrySTDinterwordspacing}{\spaceskip=0pt\relax}
\providecommand{\BIBentryALTinterwordstretchfactor}{4}
\providecommand{\BIBentryALTinterwordspacing}{\spaceskip=\fontdimen2\font plus
\BIBentryALTinterwordstretchfactor\fontdimen3\font minus \fontdimen4\font\relax}
\providecommand{\BIBforeignlanguage}[2]{{%
\expandafter\ifx\csname l@#1\endcsname\relax
\typeout{** WARNING: IEEEtran.bst: No hyphenation pattern has been}%
\typeout{** loaded for the language `#1'. Using the pattern for}%
\typeout{** the default language instead.}%
\else
\language=\csname l@#1\endcsname
\fi
#2}}
\providecommand{\BIBdecl}{\relax}
\BIBdecl

\bibitem{kim2023toward}
M.~Kim, H.~Yetkin, D.~J. Stilwell, J.~Jimenez, S.~Shrestha, and N.~Stark, ``Toward optimal placement of spatial sensors to detect poisson-distributed targets,'' \emph{IEEE Access}, 2023.

\bibitem{diggle2006spatio}
P.~J. Diggle, ``Spatio-temporal point processes, partial likelihood, foot and mouth disease,'' \emph{Statistical methods in medical research}, vol.~15, no.~4, pp. 325--336, 2006.

\bibitem{moller1998log}
J.~M{\o}ller, A.~R. Syversveen, and R.~P. Waagepetersen, ``Log {G}aussian {C}ox processes,'' \emph{Scandinavian Journal of Statistics}, vol.~25, no.~3, pp. 451--482, 1998.

\bibitem{jullum2020estimating}
M.~Jullum, T.~Thorarinsdottir, and F.~E. Bachl, ``Estimating seal pup production in the {G}reenland {S}ea by using {B}ayesian hierarchical modelling,'' \emph{Journal of the Royal Statistical Society: Series C (Applied Statistics)}, vol.~69, no.~2, pp. 327--352, 2020.

\bibitem{shirota2017space}
S.~Shirota and A.~E. Gelfand, ``Space and circular time log {G}aussian {C}ox processes with application to crime event data,'' \emph{The Annals of Applied Statistics}, pp. 481--503, 2017.

\bibitem{kim2023role}
M.~Kim, H.~Yetkin, D.~J. Stilwell, and J.~Jimenez, ``On the role of uncertainty in poisson target models used for placement of spatial sensors,'' in \emph{Ocean Sensing and Monitoring XV}, vol. 12543.\hskip 1em plus 0.5em minus 0.4em\relax SPIE, 2023, pp. 65--77.

\bibitem{kim2024SysCon}
M.~Kim, D.~J. Stilwell, H.~Yetkin, and J.~Jimenez, ``Near-optimal sensor placement for detecting stochastic target trajectories in barrier coverage systems,'' \emph{IEEE SysCon}, 2025.

\bibitem{diggle2005point}
P.~Diggle, B.~Rowlingson, and T.-l. Su, ``Point process methodology for on-line spatio-temporal disease surveillance,'' \emph{Environmetrics: The Official Journal of the International Environmetrics Society}, vol.~16, no.~5, pp. 423--434, 2005.

\bibitem{rue2009approximate}
H.~Rue, S.~Martino, and N.~Chopin, ``Approximate {B}ayesian inference for latent {G}aussian models by using integrated nested {L}aplace approximations,'' \emph{Journal of the Royal Statistical Society: Series b (statistical methodology)}, vol.~71, no.~2, pp. 319--392, 2009.

\bibitem{lindgren2015bayesian}
F.~Lindgren and H.~Rue, ``{B}ayesian spatial modelling with {R-INLA},'' \emph{Journal of statistical software}, vol.~63, pp. 1--25, 2015.

\bibitem{bachl2019inlabru}
F.~E. Bachl, F.~Lindgren, D.~L. Borchers, and J.~B. Illian, ``inlabru: an {R} package for {B}ayesian spatial modelling from ecological survey data,'' \emph{Methods in Ecology and Evolution}, vol.~10, no.~6, pp. 760--766, 2019.

\bibitem{marinecadastre.gov}
\emph{\em Bureau of Ocean Energy Management (BOEM) and National Oceanic and Atmospheric Administration (NOAA). MarineCadastre.gov. {2020 Vessel Traffic Data}. Retrieved: September 5th 2022. from marinecadastre.gov/data.}\hskip 1em plus 0.5em minus 0.4em\relax marinecadastre.gov, 2020.

\bibitem{rpackagedocumentation_2019}
\BIBentryALTinterwordspacing
``Matern {SPDE} model object with {PC} prior for {INLA},'' Dec 2019. [Online]. Available: \url{https://rdrr.io/github/INBO-BMK/INLA/man/inla.spde2.pcmatern.html}
\BIBentrySTDinterwordspacing

\end{thebibliography}

% \bibliographystyle{IEEEtran}
% \bibliography{IEEEabrv, main.bib}
% \bibliography{IEEEabrv,bib_files}

\end{document}